\documentclass[journal]{IEEEtran}

\usepackage{graphicx}
\usepackage{amsmath,amssymb} 
\usepackage{caption}

\usepackage{url}

\usepackage{algorithm}
\usepackage{algpseudocode}

\errorcontextlines\maxdimen

\makeatletter
\newcommand*{\algrule}[1][\algorithmicindent]{\makebox[#1][l]{\hspace*{.5em}\thealgruleextra\vrule height \thealgruleheight depth \thealgruledepth}}%
\newcommand*{\thealgruleextra}{}
\newcommand*{\thealgruleheight}{.75\baselineskip}
\newcommand*{\thealgruledepth}{.25\baselineskip}

\newcount\ALG@printindent@tempcnta
\def\ALG@printindent{%
	\ifnum \theALG@nested>0
	\ifx\ALG@text\ALG@x@notext
	\else
	\unskip
	\addvspace{-1pt}
	\ALG@printindent@tempcnta=1
	\loop
	\algrule[\csname ALG@ind@\the\ALG@printindent@tempcnta\endcsname]%
	\advance \ALG@printindent@tempcnta 1
	\ifnum \ALG@printindent@tempcnta<\numexpr\theALG@nested+1\relax
	\repeat
	\fi
	\fi
}%
\usepackage{etoolbox}
\patchcmd{\ALG@doentity}{\noindent\hskip\ALG@tlm}{\ALG@printindent}{}{\errmessage{failed to patch}}
\makeatother

\newbox\statebox
\newcommand{\myState}[1]{%
	\setbox\statebox=\vbox{#1}%
	\edef\thealgruleheight{\dimexpr \the\ht\statebox+1pt\relax}%
	\edef\thealgruledepth{\dimexpr \the\dp\statebox+1pt\relax}%
	\ifdim\thealgruleheight<.75\baselineskip
	\def\thealgruleheight{\dimexpr .75\baselineskip+1pt\relax}%
	\fi
	\ifdim\thealgruledepth<.25\baselineskip
	\def\thealgruledepth{\dimexpr .25\baselineskip+1pt\relax}%
	\fi
	\State #1%
	\def\thealgruleheight{\dimexpr .75\baselineskip+1pt\relax}%
	\def\thealgruledepth{\dimexpr .25\baselineskip+1pt\relax}%
}

\begin{document}

\title{A Systematic Approach for Cross-source Point Cloud Registration by Preserving Macro and Micro Structures}

\author{Xiaoshui~Huang, 
        Jian~Zhang, 
        Lixin~Fan, 
        Qiang~Wu, 
        Chun~Yuan
\thanks{Xiaoshui Huang, Jian Zhang and Qiang Wu are with the Global Big Data Technologies Centre (GBDTC), School of Computing and Communication. University of Technology Sydney,  Australia. (Email: Xiaoshui.Huang@student.uts.edu.au, Jian.Zhang@uts.edu.au and Qiang.Wu@uts.edu.au). }
\thanks{Lixin Fan is affiliated with Nokia Technologies, Tampere, Finland (Email: lixin.fan@nokia.com). }%
\thanks{Chun Yuan is with the Graduate School of Shenzhen, Tsinghua University, China (yuanc@sz.tsing.edu.cn). } }


\maketitle

\begin{abstract}
We propose a systematic approach for registering cross-source point clouds. The compelling need for cross-source point cloud registration is motivated by the rapid development of a variety of 3D sensing techniques, but many existing registration methods face critical challenges as a result of the large variations in cross-source point clouds. This paper therefore illustrates a novel registration method which successfully aligns two cross-source point clouds in the presence of significant missing data, large variations in point density, scale difference and so on. 
The robustness of the method is attributed to the extraction of macro and micro structures. Our work has three main contributions: (1)  a systematic pipeline to deal with cross-source point cloud registration; (2)  a graph construction method to
maintain macro and micro structures; (3) a new graph matching method is proposed which considers the global geometric constraint to robustly register these variable graphs. Compared to most of the related methods, the experiments show that the proposed method successfully registers in cross-source datasets, while other methods have difficulty achieving satisfactory results. The proposed method also shows great ability in same-source datasets.

\end{abstract}

\section{Introduction}
\label{introduction}
There is currently a wide diversity of techniques for obtaining point clouds (e.g. Kinect, Lidar, range cameras, structure from motion (SFM) and simultaneous localization and mapping (SLAM)). Their registration is a long standing and difficult challenge in computer vision, computer graphics, robotics, and medical applications. Because a point cloud usually contains tens of thousands, or millions, of points in each scene, it is much complex and difficult than the point set registration problem, which always processes less than 1000 points in each scene. When the point cloud coming from different kinds of sensors, the registration problem becomes much more difficulty because of the disparate sensing mechanisms. However, current researches mainly focus on reporting less than 1000 point set registration \cite{cpd}\cite{2011gmm}\cite{FGM}\cite{jrmpc}\cite{ma2013robust}. Unlike these previous methods, we propose a method in this paper to deal with the cross-source point cloud registration problem, which is a generalization of complex point cloud registration. The existing point cloud registration methods can be categorized into two aspects: same-source and cross-source.

In terms of same-source point cloud registration, existing methods can be divided into two categories: direct methods and transformed methods. Direct methods usually minimize the distance between pair-wised points or features \cite{t4,icp,rpm,cmvs,huber2003fully,kinectfusion,t5,torsello2011multiview,goicp,4pcs,super4pcs}. Transformed methods usually transform 3D points from Euclidean space to other models and convert the registration problem into a model correspondence problem \cite{cpd,2011gmm,2008gmm,deng2014riemannian,torki2010putting,cleju2007stochastic}.

There is a single paper about cross-source point cloud matching/registration \cite{peng2014street}, but its registration is executed using conventional iterative closest point (ICP) \cite{icp} and many assumptions are made, including removing sparse outliers and manually selecting the dense point regions. A 4-Points Congruent Sets-based method (4PCS) shows elements of experiments that deal with cross-source problems \cite{super4pcs}, although such 4PCS-based methods are sub-optimal in the direction of slippage as a result of operating at a point level \cite{gelfand2004shape,4pcs}. 

\begin{figure}
	\centering
	\includegraphics[height=4.1cm,width=8.2cm]{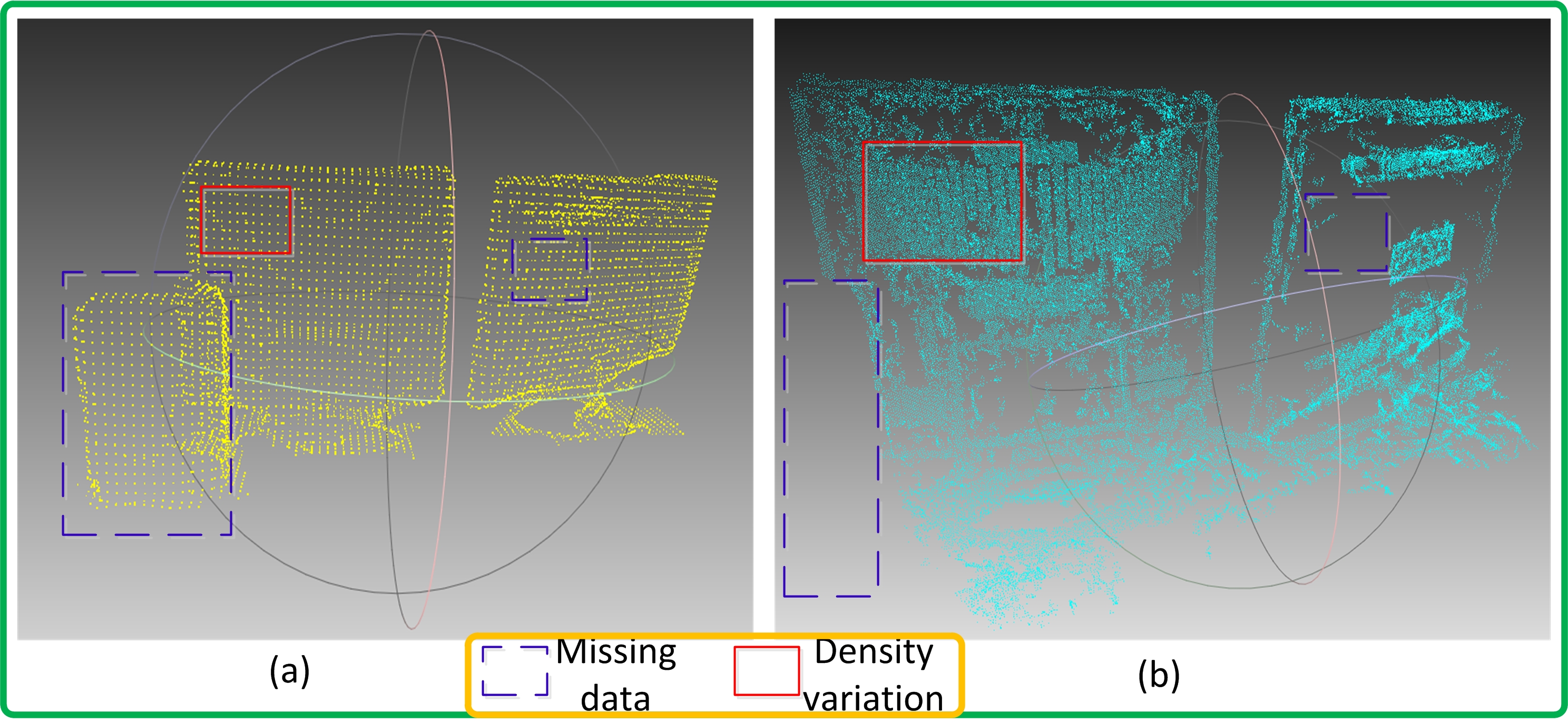}
	\caption{Examples of cross-source point clouds coming from different sensors. 3D points are produced by two sensors. (a) Kinect point cloud, (b) Point cloud reconstructed from 2D images captured by an RGB camera.}
	\label{f1}
\end{figure}

The fact that point clouds come from different kinds of sensors (e.g. SFM with mobile phones, Kinect, range cameras and Lidar) present many challenges and the existing related methods have many limitations. There is a paucity of research in the literature on this issue. In real applications, however, multiple types of sensors have much greater ability than single sensors. For example, SLAM \cite{huang2016real} constructs real-time complete depth and convert the depth to a point cloud, SFM uses images captured by RGB cameras to create point clouds for urban scenes \cite{musialski2013survey} and heritage objects \cite{manferdini2012methodology}. Other devices, such as Kinect and Lidar, offer effective ways of producing standard point cloud datasets. With the development of new technology, there are increasing means of sensing 3D point clouds describing the same objects or scenes. Registering these datasets will have great value in cultural heritage protection, city development and technology. Nevertheless, cross-source point cloud registration presents a challenge. 

Figure 1 shows two cross-source point clouds with two monitors and audio equipment, which illustrates the difficulties confronting a robust registration method:

(1) \textit{Varying densities}:  large variations in the density of cross-source point clouds often lead to the failure of existing registration methods.  

(2) \textit{Missing data}: data is often missing when the same objects have different reflection or non-reflection in various types of sensors as a result of the imaging mechanisms of different sensing techniques. For instance, this problem is pronounced for point clouds created by SFM which is unable to generate points in uniform image regions. 

(3) \textit{Large variations in scale and rotation angles}:  even though a registration method is supposed to recover scale and rotation angles, exceedingly large variations in scale and angle are often outside the capture zones of many existing methods (see Figure 7 for an example). 

As demonstrated in our experiments, these combined challenges make cross-source point cloud registration difficult, and many existing methods fail in such adverse scenarios. 

Despite the large variations in cross-source point clouds, our human vision system seems able to align them effortlessly with high accuracy. This is probably due to the fact that humans exploit the similarities between the \textit{structures of two cross-source point clouds} instead of the detailed points. Motivated by this insight, a method is proposed to extract and describe the macro structure (e.g. the global outline of objects) and the micro structure (e.g. voxels and segments) of point clouds. These macro and micro structures act like a net to robustly describe the invariant components of cross-source point clouds, and graph theory is a strong tool for preserving these structures from a mathematical viewpoint. 
A structure preserved representation method that ignores local point cloud details is proposed to deal with missing data and varying density. A scale normalization method is proposed to deal with the scale problem, and a systematic approach using these two methods is proposed to deal with all cross-source point cloud registration problems.

To the best of our knowledge, this is the first time a method has been proposed that successfully registers two cross-source point clouds in adverse scenarios. The proposed approach preserves the structure properties well by firstly, extracting reliable macro and micro structures to be robust to large noise, outliers and some of the missing data; secondly, integrating the point cloud structures as graphs and describing them; thirdly, finding the optimal graph-matching solution, and lastly, refining the solution with 3D RANSAC (Random Sample Consensus) to remove outliers and ICP to finalize the outlier-free registration. 

The contributions of this work are (1) a feasible structure-based framework to deal with the cross-source point cloud registration problem; (2) a new graph construction method to practically integrate macro and micro structures as a graph and robustly describe these structures; and (3) a new iteration method to solve the graph matching problem taking the global geometrical constraint into consideration.

\section{Related work}

Same-source point clouds are captured from the same kinds of sensors (e.g. all captured by Kinect), while cross-sources are captured from different kinds of sensors (e.g. one by Kinect, another by an RGB camera). In this section, the related methods are reviewed in terms of their ability to deal with the three challenges of cross-source point cloud registration. As noted in Section \ref{introduction}, existing methods can be categorized as: direct methods and transformed methods.

\subsection{Direct methods} 

Direct point set registration methods usually minimize the Euclidean distance between nearby points. The most popular approach is the ICP \cite{icp} algorithm, which alternates between estimating the point correspondence and estimating the transformation matrix for a given correspondence \cite{kinectfusion,huber2003fully,torsello2011multiview}. 
The vanilla ICP method \cite{icp} relies on the assumption that all points have pairwise counterparts between two sets and are very sensitive to a given initialization. The method is widely used in same-source and cross-source registration. Iterative non-rigid point set matching has improved ICP by incorporating outlier detection in the iterative correspondence estimation steps \cite{rpm}. The above methods are all heuristic methods, hence they cannot guarantee the global optimality of the solutions. Go-ICP \cite{goicp} provides a globally optimal solution to ICP in 3D Euclidean registration, which combines ICP with a branch-and-bound (BnB) scheme. Similarly, GOGMA \cite{gogma} combines Gaussian mixture model (GMM) with a BnB scheme. These global optimal methods are sensitive to scale problem. A roots-finding technique was used in \cite{ho2009algebraic} for affine invariant point set registration. The method is sensitive to outliers due to use of moments. To tackle outlier sensitive problem in \cite{ho2009algebraic}, \cite{ma2013robust} proposes a method that uses an L2E estimator and ICP. The method of \cite{ma2013robust} is suitable both for 2D and 3D situations. In the 2D instance, shape context is used as descriptor and the Hungarian method is used for matching with the $\chi^2$ test statistic as the cost measure. In the 3D instance, the spin image can be used as a feature descriptor, where the local similarity is measured by an improved correlation coefficient. \cite{ma2013robust} uses L2E, which is particularly appropriate for analyzing massive data sets when data cleaning is impractical. With the ICP refinement algorithm, this algorithm robustly estimates transformation $f$ with noise and outlier points. Also, the initial correspondences do not need to be highly accurate. The experimental results show that this algorithm has good performance under deformation, occlusion, rotation, noise and outliers. However, experiments have only been conducted on the algorithm in same-source situation; in the cross-source situation, the L2E estimator may face the problem of large outliers.  Despite these improvements to the ICP method, the direct registration approaches above are intrinsically sensitive to missing data, large variations in point density, and scale differences,  thus rendering them useless for cross-source point cloud registration (see the experimental results in Section \ref{experiment} for examples). 

In contrast to these ICP-based methods, registration amounts to solving a global problem to find the best aligning rigid transform over the 6DOF space of all possible rigid transforms comprised of translations and rotations when scan pairs start in arbitrary initial poses. Since aligning rigid transforms are uniquely determined by three pairs of (non-degenerate) corresponding points, one popular strategy is to invoke RANSAC \cite{ransac} to find the aligning triplets of point pairs \cite{chen1999ransac}. This approach, however, regularly degrades to its worst case $O(n^3)$ complexity in the number $n$ of data samples in presence of partial matching with low overlap. Various alternatives to RANSAC have been proposed to counter the cubic complexity, such as hierarchical representation in the normal space \cite{diez2012hierarchical}; super-symmetric tensors to represent the constraints between the tuples \cite{cheng2013supermatching} ; stochastic non-linear optimization to reduce the distance between scan pairs \cite{papazov2011stochastic}; branch-and-bound using pairwise distance invariants \cite{gelfand2005robust}; or evolutionary game theoretic matching \cite{albarelli2010loosely,rodola2013scale}. However, these methods are all sensitive to missing data.

Following the concept of RANSAC, another kind of method is 4PCS \cite{4pcs}, which uses a randomized alignment approach and the idea of planar congruent sets to compute optimal global rigid transformation. The 4PCS method is widely used and has been extended to take into account uniform scale variations \cite{scale4pcs}. However, these methods  have a complexity of $O(n*2+k)$ where $n$ denotes the size of the point clouds and $k$ is the set of candidate congruent 4-points. It has great limitations when point numbers are large. To remove the  quadratic complexity of the original 4PCS, \cite{super4pcs} extends it to a fast algorithm with only linear computation time needed. This method reports the points or spheres in $R^3$ and uses a smart index to quickly find the matched plane in all candidate congruent 4-points planes. One cross-source point cloud registration experiment is reported in \cite{super4pcs}. However, these methods have many limitations due to their point-level operation. They may easily be sub-optimal when computing their transformation relations. The varying density of the cross-source problem makes the performance of the 4PCS-based method even worse.

Although these direct methods show some ability in addressing elements of the cross-source problem, none of them can deal with the complete cross-source problem. In this paper, a novel method is proposed to robustly deal with the entire cross-source problem. The method extracting and combining macro and micro structures is robust to large variations in density, noise and outliers. In addition, the enhanced graph matching globally registers two structures. Lastly, a scale normalization step is used to eliminate most of the scale variation.

\subsection{Transformed methods}

One of the mathematical tools typically used for registration is Mutual Information (MI), which catches the non-linear correlations between the point clouds and the geometric properties of the target surface. The authors in \cite{sinha2002cortical} use ICP and mutual information (MI) to  build one-to-one correspondence between an magnetic resonance (MR) surface and laser-scanned cortical surface; however, this method is highly dependent on initialization and overlap rate. The work in \cite{pandey2012toward} registers unstructured 3D point clouds by using K-means to form a set of codewords and using an estimator to optimize the MI value to obtain the final rigid relations. Cross correlation of the horizontal cross section images of the two point clouds is used in \cite{moussa2015automatic} to coarsely register the point clouds, and ICP is then used to refine the coarse results. These MI-based methods perform poorly when data is missing because it make the MI of two point clouds originally not the same.

Another type of transformed method is the feature-based method, which extracts features from 3D point clouds and transforms the point cloud registration Euclidean space into feature space. Typical 3D feature extraction methods  \footnote{There is a tutorial about 3D features. \url{http://robotica.unileon.es/index.php/PCL/OpenNI_tutorial_4:_3D_object_recognition_(descriptors)}} are FPFH \cite{rusu2009fast}, ESF \cite{esf}, Spin image \cite{johnson1997spin} and SHOT \cite{tombari2010unique}. These feature-based methods produce exciting results on same-source point clouds. However, it is very difficult to reliably extract similar features from cross-source point clouds, and these methods always fail in this situation. This is because these features may original perceive large discrepancy and cannot used for registration.

Torki and Elgammal \cite{torki2010putting} use local features in images to learn manifold symbol. The authors first learn a feature embedding representation that contains the spatial structure of the features as well as the local appearance similarity. The out-of-sample method is then used to embed the features from new images. Similarly, Yuan \cite{deng2014riemannian} transforms every point in a point clouds into a shape representation, in order to cast the problem of point sets matching as a shape registration problem, which is the Schrodinger distance transform (SDT) representation. The problem is then transformed into solving a static Schrodinger equation in place of the consistent static Hamilton-Jacobi equation in the setting. The SDT representation is an analytic expression which can be normalized to have unit L2 norm in accordance with theoretical physics literature. The outline of this method is "points set"$->$ "SDTs"$->$"minimize the geodesic distance". 

\begin{figure*}[ht]
	\centering
	\includegraphics[height=5.5cm,width=122mm]{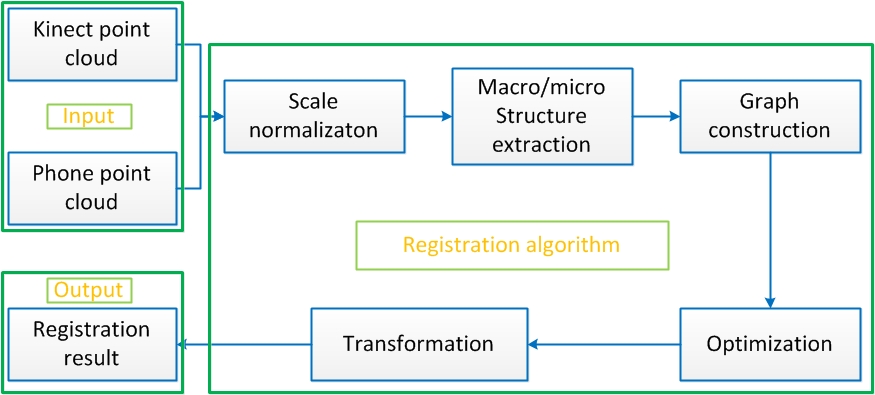}
	
	\caption{Overall system workflow.}
	\label{f2}
\end{figure*}
Related to point cloud registration, another kind of methods is GMM-based methods.  To deal with the noise and outliers existing in the point sets registration problem, Bing et al. \cite{2011gmm} proposed a method in which point clouds were represented as Gaussian Mixture Models (GMM) and, went on to solve the registration problem by minimizing the statistical discrepancies between corresponding GMMs. This approach can be used for both rigid and non-rigid point cloud registration, and has demonstrated its ability to deal with noise and outliers to some extent. Georgios et al. \cite{cpd} introduced a motion drift idea into the GMM framework and achieved good results on rigid and non-rigid point set registration. A solution to the GMM-based approach by recasting registration as a clustering problem was proposed in  \cite{jrmpc}. However, there are an increasing number of GMM models to robustly represent point clouds. When the point number increases to tens of thousands or millions, these methods are impractical in terms of both computational and memory cost. On the other hand, the GMMs depicting two point clouds are shown a lot of difference when there is missing data and large noise and outliers variations in cross-source point clouds, which makes the registration inaccurate or it may even fail. The experiments in Section \ref{experiment} demonstrate these approaches do not lead to satisfactory results for cross-source point cloud registration.

The above transformed methods demonstrate ability in dealing with parts of noise and outliers or density variation, but none of them can successfully address the cross-source registration problem, which comprises issues of scale, density variation, noise and outliers and missing data. In this paper, we aim to address this tough cross-source problem. Motivated by our human registration process,  a structure-based framework is proposed to robustly register two cross-source point clouds.

The remaining sections of this paper are organized as follows: Section \ref{SmacroAndmicro} describes the proposed macro and micro structure representation; Section \ref{registrationAlgorithm} describes the proposed registration method based on our novel concept; and Section \ref{experiment} describes the experiments, and Section \ref{conclusion} concludes the paper.

\section{Macro and micro structure representation}
\label{SmacroAndmicro}

As mentioned in Section \ref{introduction}, the significant challenges for 3D cross-source point cloud registration are the large variations in density, missing data, scale and angle between two point clouds. To address these variations, we define two structures, known as macro and micro structures, to describe the point clouds based on our observations. In our work, we extract structures from the cross-source point clouds and use these structures to indirectly register cross-source point clouds, instead of try to deal with these difficult changing points directly. Similar to our human ability, these structures robustly describe the global and local invariance of the cross-source point clouds, even though there are many variations in relation to these point clouds.    

\begin{figure}[h]
	\centering
	\includegraphics[height=4cm,width=80mm]{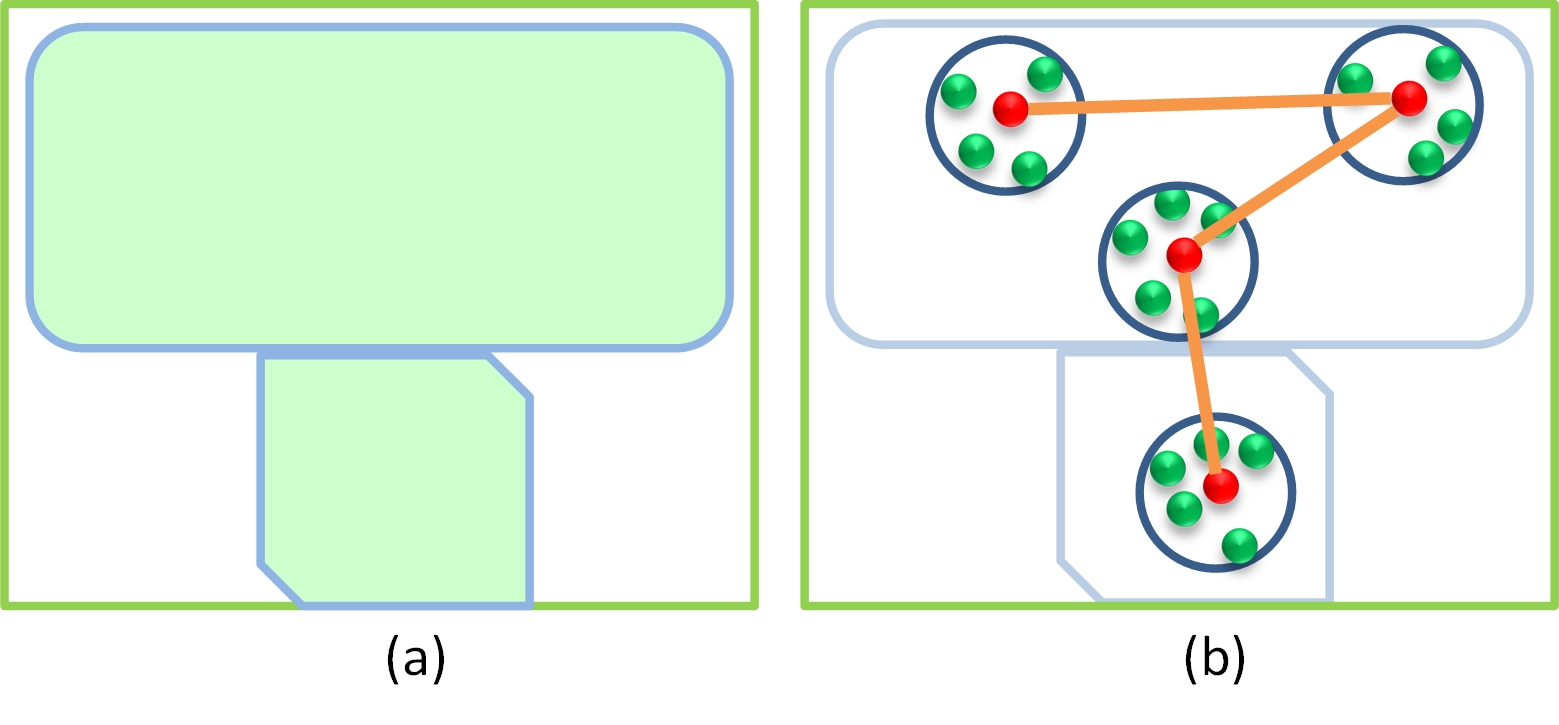}
	\caption{Schematic diagram of macro and micro structures.}
	\label{miAndma}
\end{figure}

The macro structure is the overall outline or large-scale structure of an object or scene. It is important to note that it represents the global properties of the structure, such as the boundary outline, the contour and the shape, but not the global light, global color or global material. Figure \ref{miAndma}(a) illustrates that the rectangle above the square (the blue outline) is the macro structure. When humans judge whether two objects are similar, they usually first consider the macro structure, and an overall alignment is obtained on this basis. We define a micro structure to work alongside the macro structure. The micro structure is defined as a small scale structure, such as a stable cell or part of the object or scene. It is a local property that describes the internal details of the object or scene. In our work, the micro structure consists of a 3D region, such as a super voxel in 3D point clouds. Figure \ref{miAndma}(b) illustrates that, super voxels contain points with the same properties of 3D spatial geometry. We use these micro and macro structures to iteratively obtain the corresponding relations between two point clouds.

\section{Registration algorithm}
\label{registrationAlgorithm}

In this section, we describe the registration method based on the proposed macro and micro structure theory and describe the components that make up our system. Figure \ref{f2} provides an overview of our method in block form. It comprises the following five components:

\emph{\textbf{Step 1, Scale normalization:}} The pre-processing stage. Two cross-source point clouds, which come from different sensors, are normalized to the same scale. The details are given in Section \ref{scale}.

\emph{\textbf{Step 2, Macro/micro Structure Extraction:}} The main novelty of this stage is the reliable extraction of the structure from large variable cross-source point clouds, which is robust to most cross-source problems. These point clouds are segmented into many super voxels, using their 3D geometric properties, and the statistical property of each super voxel is used for its robust to local variations. These super voxels are integrated as the macro structure and the statistical property of each super voxel becomes the micro structure, as detailed in Section \ref{voxelExtranction}. These structures are integrated in the next step. 

\emph{\textbf{Step 3, Graph construction:}} The main novelty of this stage is the combination of micro and macro structures using graphs. Although there are many variations in two cross-source point clouds, the invariant structure properties are preserved in this method. The nodes are the extracted voxels and the edges are the adjacent relations. In addition, a new similarity measure method is proposed which robustly describes these two graphs, as detailed in Section \ref{graphconstruction}. After the graph has been constructed, the registration problem is converted to a graph matching problem. An optimization method is thus needed to optimize the graph matching problem.

\emph{\textbf{Step 4, Optimization:}} The novelty of this stage is the proposal of an enhanced optimization method. Factorized graph matching \cite{FGM} is an optimization algorithm that optimizes graph matching at a constant time and is less prone to local optimization. To better suit to our problem and to pursue global optimal, we consider the geometry constraints in our optimization as detailed in Section \ref{optimization}. With this matching result, further refinement is needed.

\emph{\textbf{Step 5, Transformation estimation:}} Transformation matrix computation stage. RANSAC is performed to first remove outliers, following which ICP refines the initial matching from the graph matching, as detailed in Section \ref{transformation}.

\subsection{Scale normalization}
\label{scale}
The two point clouds come from different sensors and therefore have different scales. To remove scale variation, we conduct scale normalization before the super voxel extraction step. Previously, the scale was normalized by manual measurements in the real world and these two point clouds were calibrated, but although manual measurement is accurate, it is sometimes difficult. We propose an automatic method to estimate the scale without the need for manual work. To achieve this goal, we first compute the mean distance of two 3D points and then compute the scale by comparing these two means as follows:

\begin{eqnarray}
	\begin{aligned}
		scale=\frac{max||P_i-\overline P||_2}{max||Q_i-\overline Q||_2}
	\end{aligned}
\end{eqnarray}
where $\overline{P}=(\sum\limits_{i=1}^{N}{P_i})/N$ and $\overline{Q}=(\sum\limits_{j=1}^{M}{Q_j})/M$ 

We use this scale to transform other point clouds and remove the scale difference in cross-source point clouds as far as possible. Although we cannot deal with the scale problem completely, the results of this stage are sufficient for the graph matching stage since most of the scale difference is eliminated. After the scale difference has been removed, the voxels can be extracted.

\subsection{Macro/micro Structure extraction}
\label{voxelExtranction}
Due to the large variations in cross-source point clouds, a method is needed to extract the invariable components. Figure \ref{f1} shows that even though the two cross-source point clouds have many variations, the structure can still be recognized. For these cross-source point clouds, therefore, the focus is on the structure information rather than the detailed information, since the latter is full of noise, outliers and different densities.

We are motivated by the idea of cluster, where points with the same property are clustered together. As shown in Figure 1, humans have the ability to register these monitors at first glance. This is because the macro structure information remains in the cross-source data and when humans conduct the registration work, they are not concerned with information detail (e.g. the location of a point). However, if we want to accurately register these two point clouds, macro structure information alone is insufficient, and micro structure information is also needed. Hence, to develop an intelligent registration algorithm, we need a method that will retain the common macro and micro structure information and ensure it is robust to varying densities and missing data. 

\begin{figure}[ht]
	\centering
	\includegraphics[height=5cm,width=80mm]{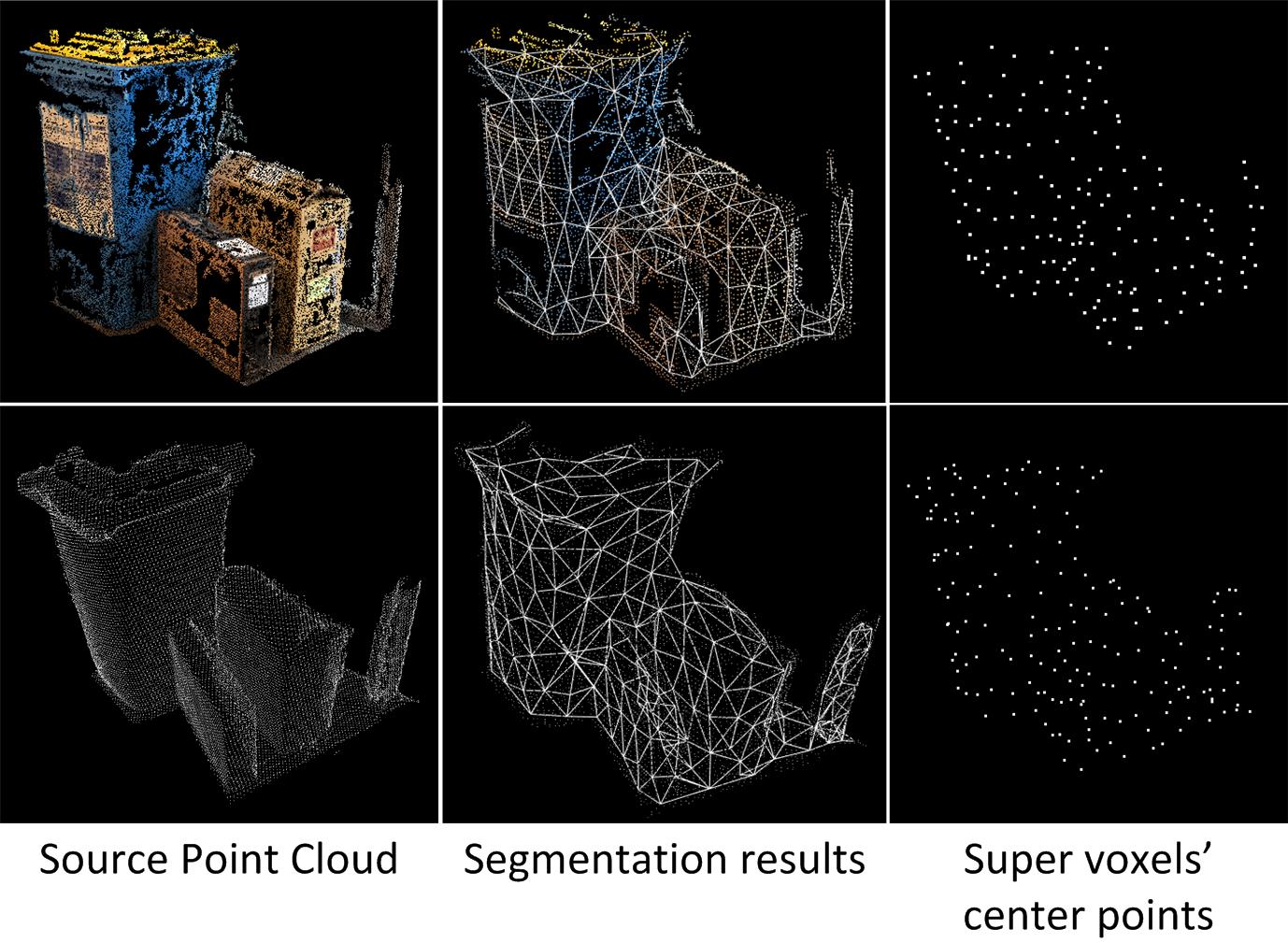}
	\caption{Results of macro/micro structure extraction. The first column is the source point clouds of SFM (above) and KinectFusion (bottom); the second column is the segmentation results and the connection relationship; the third column is the segmented super voxels' central points.}
	\label{superpixel}
\end{figure}

To fulfill this goal, we improve the recently developed segmentation method \cite{svoxel} to segment the two point clouds into many super voxels and extract the direct adjacency graph of these voxels. As the segmentation method adheres to object boundaries while remaining efficient by only using the 3D geometric property, it obtains robust results for two point clouds, regardless of different density, angle, noise and missing data (see the third column of Figure \ref{superpixel}). Figure \ref{superpixel} shows that the center of the segmented super voxels deals with much of the noise, density and missing data problem. Unlike \cite{svoxel}, we do not flow back at the extraction of each edge in the adjacency graph extraction step, which means that the direction information is considered in our new adjacency graph. This is because in the following optimization step (Section \ref{optimization}), direct graph matching achieves more robust results than indirect graph matching \cite{dfm}. This revision is a key element to ensure that these extracted voxels are correctly and robustly registered. At the same time, the ESF descriptor \cite{esf} for each voxel is extracted to describe the statistical property as a local structure.  Based on the definition of macro and micro structures, therefore, each segmented super voxel is a micro structure and the whole of the adjacency graph and voxel centers are macro structures. After these structures have been extracted, they are integrated as a graph in the graph construction stage.

\subsection{Graph construction}
\label{graphconstruction}
A new graph construction method is proposed to utilize macro and micro structures to deal with the cross-source point cloud registration problem. The new graph construction method integrates these structures and forms the registration problem into a graph matching problem. We select graph because it is a strong tool for maintaining the properties(e.g. topology) of macro structures. At the same time, the nodes and the edges of the graph are able to maintain properties of the micro structure. 

\begin{figure*}[ht]
	\centering
	\includegraphics[height=4cm,width=130mm]{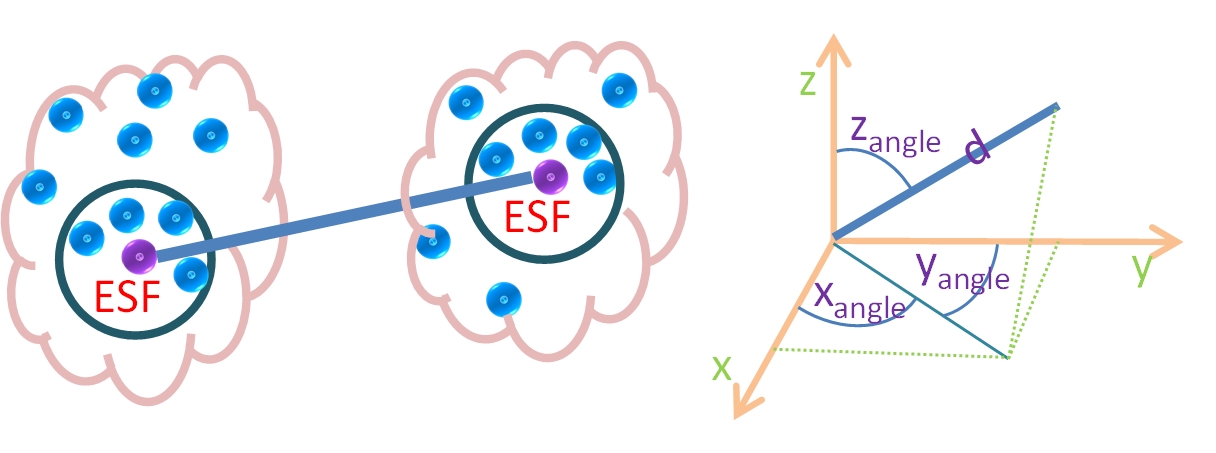}
	\caption{Schematic diagram of graph nodes and edges.}
	\label{nodeedge}
\end{figure*}

Before introducing the new method, the graph matching notations are introduced. A graph with $n$ nodes and $m$ directed edges is defined as \u{C}$ = \{P, Q, G, H\}$. $P$ and $Q$ are the features for the nodes and edges of the graph, which are defined as $P = [p_1, \cdots ,p_n] \in R^{d_p\times n} $and $Q = [q_1, \cdots ,q_m] \in R^{d_q\times m}$ respectively. For example, $p_i$ could be a SIFT descriptor or ESF descriptor extracted from the original data around the $i_th$ node and $q_i$ could be the length of the $i^{th}$ edge. $G,H \in \{0, 1\}^{n\times m}$ is a node-edge incidence matrix which describes the topology of the graph. We define $g_{ic} = h_{jc} = 1$ if the $c^{th}$ edge connects the $i^{th}$ node and the $j^{th}$ node, and zero otherwise. To perform graph matching, given a pair of graphs, we first need to define $P$ and $Q$. Next, we compute two affinity matrices, $K_p \in R^{n_1 \times n_2} $ and $K_q \in R^{m_1 \times m_2}$ to measure the similarity of each node and edge pair, then $k^p_{i_1 i_2}=\phi_p(p^1_{i_1}, p^2_{i_2})$ measures the similarity between the $i_1^{th}$ node of \u{C}$_1$ and the $i_2^{th}$ node of \u{C}$_2$, and $k^q_{c_1c_2}=\phi_q(q^1_{c_1}, q^2_{c_2})$ measures the similarity between the $c_1^{th}$ edge of \u{C}$_1$ and the $c_2^{th}$ edge of \u{C}$_2$. Only when we define these matrices correctly, can we use graph matching method.

A robust structure-based graph construction method is proposed in this paper. To robustly deal with the many variations in cross-source problem, with exception of structure extraction, a structure-retaining similarity measurement method is needed. In other words, the graph should be robustly described despite the cross-source problem. As previously discussed, humans can still register cross-source point clouds correctly by their structure. Similar to the human register's process, the graph is constructed as a expression of the relations between structures. This is another key element obtaining robust registration results. The micro structures are utilized as the node descriptors and the spatial relations of the centers of micro structures are utilized as the edge descriptors. The graph has the ability of being robust to large variations in density, angle and missing data of cross-source point clouds. Here, we describe how to design the nodes and edges of these graphs, and their similarity measurement.

\subsubsection{Graph node and similarity measurement}
To robustly represent the micro structures of point clouds, the method should be resilient to the large variations in density and missing data. We segment the super voxels of two point clouds and extract the centroid point of each super voxel. The graph node $E$ is constituted by these centroid points. To correctly match these nodes, they need to be described discriminately. Due to the cross-source problems discussed above (i.e. varying density, missing data and  large variations in scale and rotation angles), using only the coordinates of these centroid points cannot describe discriminately for node description and the original matched nodes pairs are very rare. To robust deal with the cross-source problem, we select the ESF descriptor \cite{esf} instead of using conventional nodes' coordinate because the ESF descriptor is a global descriptor that adds up the properties of the distance, angles and area of the point clouds. Using the ESF descriptor, we transform the variable Euclidean space into feature space (ESF 640). If two points come from the corresponding segments, the ESF descriptors will mostly be the same and should be matched, even though the centroid point may not perfectly match in the Euclidean space. 

The node similarity matrix $K_p$ is computed by comparing the distance between the nodes' ESF descriptors(see left hand of Figure \ref{nodeedge}). Here, the node similarity is not computed in Euclidean space but in feature space. Because ESF is a statistic and global descriptor, it has the ability to avoid the large local variations in Euclidean space and hence is more robust to the cross-source problem. The node similarity is

\begin{eqnarray}
\begin{aligned}
K_p=\overline{D}_{esf}
\end{aligned}
\end{eqnarray}

where $\overline{D}_{esf}$ is the normalized distance of two 3D points' ESF descriptors, $\overline{D}_{esf}=D_{esf}/ max(max(D_{esf}))$. $D_{esf}$ is the distance of two 3D points' ESF descriptors and $D_{esf}= \lVert {P^1_{esf}}(i) -{P^2_{esf}}(j) \lVert_2$.

\subsubsection{Graph edge and similarity measurement}
To robustly and discriminately describe the point cloud, it is necessary to build the edges accurately to reflect its macro structure. We record the adjacent relations (extracted in Section \ref{voxelExtranction}) between super voxels and use these adjacent relations as edges $Q$. The adjacent relations correctly reflect the relations of the super voxels through the boundary property. The edges need to be described discriminatingly and meaningfully to ensure they are correctly matched. We need to reiterate that humans can register these two cross-source point clouds because their structures are almost the same. We therefore need to retain the structure property of these two graphs in describing edges. Edge direction is also an important factor for the structure of the graph, in spite of the edge distance. 

In this paper, we use the spatial distance and geometric properties of these edges (see right hand of Figure \ref{nodeedge}). The Euclidean distance and Eular angles of two connected nodes are combined to construct a descriptor vector for describing the edges $Q$: $(x_{{angle}}, y_{angle}, z_{angle}, d)$, where $d= \lVert P_i -P_j \lVert_2, z_{angle}=acos(z/d), x_{angle}=acos(x/(d*sin(z_{{angle}})), y_{angle}=acos(y/(d*sin(z_{angle}))$. We compare the similarity by comparing the similarity of these descriptors and obtain $D_e$, where $D_e=\lVert Q^1_i-Q^2_j\lVert_2$. To make a more robust comparison, we normalize the descriptor $\overline{D_e}=D_e/max(D_e)$, and the edge similarity matrix $K_q$ is computed by 

\begin{eqnarray}
\begin{aligned}
K_q=\overline{D}_e
\end{aligned}
\end{eqnarray}

This is a simple means of obtaining features in 3D point clouds (Euclidean distance and Eular angles of two points). At the same, it describes the edges, taking the spatial relations and structures into consideration. Tts ability to register the cross-source point clouds will be demonstrated in the experiment section.

\subsection{Optimization}
\label{optimization}
We propose an enhanced factorized graph matching method which considers global geometry constraint to deal with the local minima problem in graph matching. Before introducing our method, we briefly review graph matching and FGM \cite{FGM}. Suppose there is a pair of graphs, \u{C}$_1 = \{P_1, Q_1, G_1\}$ and \u{C}$_2 =\{P_2, Q_2, G_2\}$. The problem of graph matching consists of finding a correspondence between the nodes of \u{C}$_1$ and \u{C}$_2$ that maximizes the following score of global consistency:

\begin{eqnarray}
\begin{aligned}
J(X)=\Sigma_{i_1i_2}x_{i_1i_2}k^p_{i_1i_2} + \Sigma_{\begin{subarray}{c}
	i_1\neq i_2, j_1\neq j_2 \\ h^1_{i_1c_1}g^1_{j_1c_1}=1 \\ h^2_{i_2c_2}g^2_{j_2c_2}=1 
	\end{subarray} } x_{i_1i_2}x_{j_1j_2}k^q_{c_1c_2}
\end{aligned}
\end{eqnarray}

where $X\in \{0,1\}^{n_1 \times n_2}$ denotes the node correspondence, for example, if $i_1^{th}$ node of \u{C}$_1$ and the $i_2^{th}$ node of \u{C}$_2$ correspond, $x_{i_1i_2}=1$ . $k^p_{i_1i_2}$ is an element of $K_p$ in $i_1^{th}$ row and $i_2^{th}$ col, $k^q_{c_1c_2}$ is an element of $K_q$ in $c_1^{th}$ row and $c_2^{th}$ col. 

It is more convenient to write $J(X)$ in a quadratic form, $x^TKx$, where $x=vec(X)\in \{0,1\}^{n_1n_2}$ is an indicator vector and $K \in R^{n_1n_2 \times n_1n_2}$ is computed as follows:

\begin{eqnarray}
\begin{aligned}
k^p_{i_1i_2j_1j_2} = \left\{
\begin{array}{lll}
{k^p_{i_1i_2}} &\text{if}\ i_1=j_1 \ \text{and} \ i_2=j_2 \\
{k^q_{c_1c_2}} &\text{if}\ i_1\neq j_1 \ \text{and}\ i_2\neq j_2\ \text{and}\ \\ & h^1_{i_1c_1}g^1_{j_1c_1}h^2_{i_2c_2}g^2_{j_2c_2}=1\\
{0} &\text{otherwise} 
\end{array}  
\right.
\end{aligned}
\end{eqnarray}

A factorized graph matching (FGM) method \cite{FGM} is used to develop an initial-free optimization scheme with no accuracy loss to address the non-convex issue. This method divides matrix $K$ into many smaller matrices. Using these smaller matrices, the graph matching optimization problem can be transformed to iteratively optimize the following non-linear problem: 

\begin{eqnarray}
\begin{aligned}
\max\limits_{X} J_\alpha(X)=(1-\alpha)J_{vex}(X)+\alpha J_{cav}(X)
\end{aligned}
\end{eqnarray}
where $J_{vex}$ and $J_{cav}$ are two relaxations in FGM \cite{FGM}.

\textbf{Enhanced factorized graph matching.} Although FGM iteratively uses a different $\alpha$ to apply the Frank-Wolfe (FW) algorithm to avoid local optimal, it still exists to some extent. To effectively deal with the local optima in FGM, we improve the algorithm by considering global geometry constraint and introduce a new iteration method to solve the new algorithm. The improved energy function is :

\begin{eqnarray}
\begin{aligned}
\begin{split}
\max \limits_{X} J_\alpha(X)=(1-\alpha)J_{vex}(X)+&\alpha J_{cav}(X)\\
+&J_{smooth}(X)
\end{split}
\end{aligned}
\end{eqnarray}

As our registration problem only has rigid rotation and translation, these rigid transformation relations always have neighbor projection errors nearby. We use this property to avoid the local minima and obtain more accurate transformation relations. We design this regulation term by considering the projection difference of neighboring correspondence points. $J_{smooth}(X)$ is defined as 

\begin{eqnarray}
\begin{aligned}
J_{smooth}(X)=-\sum_{i\in X}\sum_{j\in D}\frac{\arrowvert {\Arrowvert p_i-p_j\Arrowvert}-{\Arrowvert p_{im}-p_{jm}\Arrowvert}\arrowvert}{(n_1*n_2)}
\end{aligned}
\end{eqnarray}

where D represents connection points with point i, $p_{im}$ is the matched point of $p_i$ and $p_{jm}$ is the matched point of $p_j$. We can easily obtain these points in D by searching matrix G in the graph.

To optimize this nonlinear problem, we use FW \cite{fw}, which iteratively updates the solution of $X^*= X+\lambda Y$. Given an initial $X_0$, we update $X$ through optimal direction $Y$ and step size $\lambda$. As a smooth term needs a correspondence relation, we divide the computation of optimal direction $Y$ into two steps: (1) compute initial $Y_0$ using $J_{vex}$ and $J_{cav}$.  We compute an initial $Y_0$ by solving the Hungarian algorithm which is linear programming similar to FGM \cite{FGM}. (2) computes the final $Y$ by using $J_{vex}$, $J_{cav}$ and $J_{smooth}$. We compute the energy of the smooth terms using $Y_0$ and obtain the final $Y$ using the new energy. As the computation of $Y$ involves linear programming, adding one more computation step of $Y$ is not computationally costly. Similar to the FGM strategy, we also use 100 times iteration to discard the inferior temporary solution and compute an alternative solution using another FW step to optimize J(X). The final transformation matrix is computed in the next stage, following optimization.


\subsection{Transformation estimation}
\label{transformation}
Our goal is the registration of two cross-source point clouds. As the results of the graph matching contain a small number of outliers, we cannot use these results directly to compute the transformation matrix (used to combine two point clouds into a coordinate system).  We need to remove the outliers to obtain the final transformation matrix. We use 3D RANSAC \cite{3dransac} to remove the outliers, after which we use these inners to compute the transformation matrix and perform the transformation for the point clouds. The transformation matrix may sometimes still contain small errors, so to deal with this situation, we add an ICP step to locally refine the registration after the outlier removal process. After completing these steps, we register the two cross-source point clouds together. The pseudo code of the complete registration algorithm is shown in Algorithm \ref{IS2OSLS}.

\begin{algorithm}
   \caption{Pseudo-code of the registration algorithm. }\label{IS2OSLS}
    	\begin{algorithmic}[1]
    		\Require Cross-source  point clouds.
    		\Ensure Registration result and Transformation matrix
    		\State Scale normalization  by  Eq. (1).
    		\State Macro/micro Structure extraction.
    		\State Graph construction using Eq. (2)  and Eq. (3).
			\State Initialize X to be a doubly stochastic matrix;
    		\For{$\alpha$ = $0:0.01:1$}
    		\For{$nIt$ = $1:100$}
    		\State Compute {$J_{vex}$} and {$J_{cav}$} from $X_0$
    		\State Compute $Y_0$ using {$J_{vex}$} and {$J_{cav}$}
    		\State Compute $J_{smooth}$ using $Y_0$ as Eq. (8)
    		\State Compute Y using {$J_{vex}$}, {$J_{cav}$} and $J_{smooth}$
    		\State Compute the update direction $Y=Y-X_0$
    		\State Compute update step $\lambda$
    		\State Compute the updated  X and set $X_0=X$    		
    		\EndFor
    		\EndFor
    		
    		\State Transformation estimation.

    	\end{algorithmic}
\end{algorithm}

\section{Experiments}
\label{experiment}

The proposed method provides a solution to the cross-source point cloud registration problem. In this section, we conduct comparative experiments with many state-of-the-art registration methods: first, we compare the performance of the method on same-source datasets, and then conduct thoroughful experiments on challenging cross-source datasets.

\subsection{Experimental setup}
For comparison purposes, we select the representative 3D registration algorithms ICP \cite{icp}, Go-ICP \cite{goicp}, 4PCS \cite{4pcs}, super-4PCS \cite{super4pcs}, TPS-RPM \cite{rpm}, GMMReg \cite{2011gmm}, CPD \cite{cpd} and JP-MPC \cite{jrmpc} as our compared methods. Experiments cannot be conducted on a large number of point cloud registrations using TPS-RPM and JR-MPC due to the memory cost, so to make a fair and reasonable comparison, we downsample the original point cloud and let the number of points be approximately 2000.

For the same-source database, we conduct a quantitative evaluation experiment with the 3D models "Bunny", "Lucy" and "Armadillo" from the Stanford 3D scanning repository\footnote{https://graphics.stanford.edu/data/3Dscanrep/3Dscanrep.html}. We only consider points with positive z coordinates. For each view, following \cite{jrmpc}, the original models are rotated in the xz-plane and the points with negative $z$ coordinates are rejected. In this way, only a part of the object is viewed in each set; the point sets do not fully overlap, and the extent of the overlap depends on the rotation angle, as in real scenarios. 

There are three types of cross-source database: 

Database A: KinectFusion and Phones' RGB camera. We build a database with four sets of cross-source objects, which are typical examples of the different properties of cross-source point clouds. We use KinectFusion to build one source, and use VSFM to build another source for images which are captured by IPhone 6S Plus. As KinectFusion uses a physical device to capture 3D points, it can usually obtain dense and uniform point clouds on an object's surface. However, VSFM is a method by which 3D point clouds are built from 2D images. It uses keypoints to initially build highly accurate 3D points and uses CMVC \cite{cmvs} to build more dense 3D points. These two sources are typical examples  of cross-source problems, as previously discussed. 

\begin{figure}[ht]
	\centering
	\includegraphics[height=4.5cm,width=8cm]{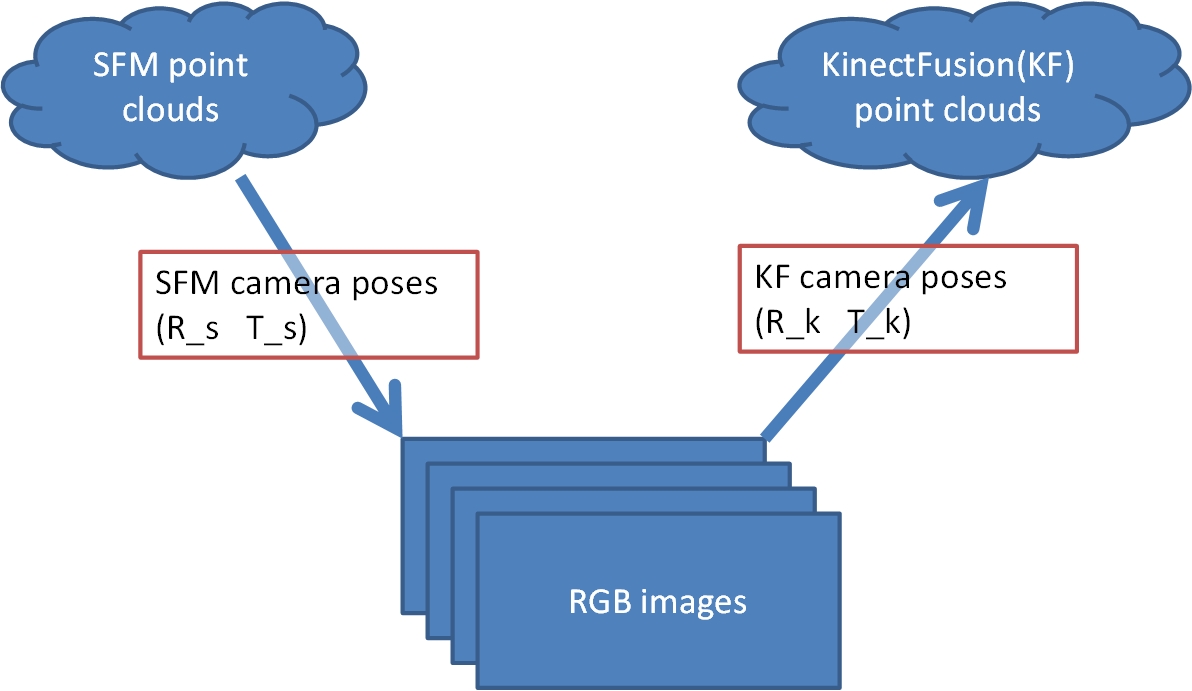}
	\caption{Theory of database B build-up. VSFM point cloud is back-projected into image coordinate system and is re-projected into KinectFusion coordinate system}
	\label{database}
\end{figure}

\begin{figure*}[ht]
	\centering
	\includegraphics[height=6cm,width=140mm]{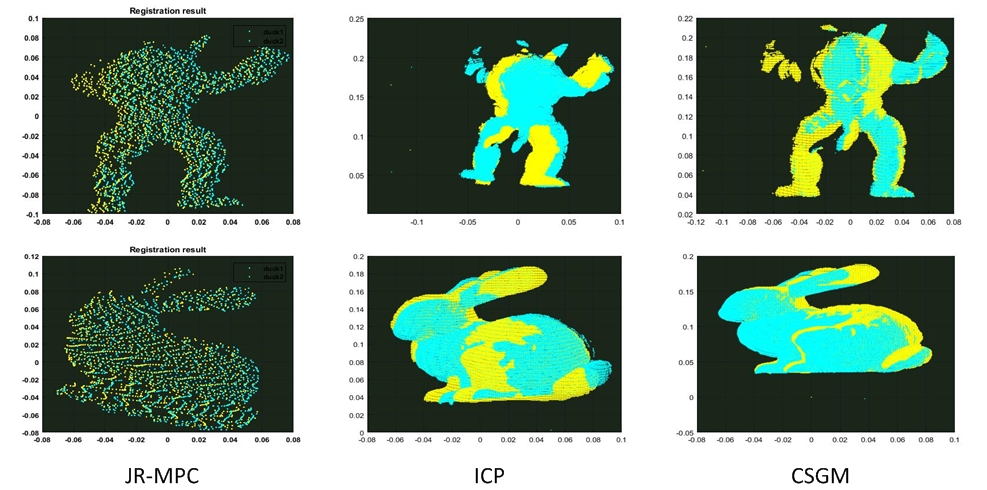}
	\caption{ Two point clouds registration results on same-source datasets. }
	\label{f7}
\end{figure*}

Database B: KinectFusion and KinectFusion's RGB camera. We build the database in the following steps: Step 1, the original KinectFusion SDK \footnote{https://www.microsoft.com/en-au/download/details.aspx?id=40276} is revised to output the image sequence and camera pose of each image when capturing KinectFusion point clouds. Step 2, another point cloud is computed using these images and VSFM. A set of camera poses is computed using VSFM. As these two cross-source point clouds come from the same set of image sequences, the camera poses of KinectFusion and VSFM should be the same. Using this theory, a cross-source point cloud database is produced. The theory is illustrated shown in Figure \ref{database}. The VSFM point cloud is back-projected into the image coordinate system and then re-projected into the KinectFusion coordinate system. To avoid the inaccuracy of camera pose computation in VSFM and KinectFusion, we consider many poses whose reprojection error is less than $\sigma$ ($\sigma$=0.5), and use these camera pose center points and the least-squares method to compute the final rigid transformation between these two camera center points. The rigid transformation matrix is built on critical prior information and can therefore be used as ground-truth. These benchmark data contain 13 datasets and can be used to perform quantitative evaluation for cross-source point cloud registration.

\begin{figure}[ht]
	\centering
	\includegraphics[height=4.5cm,width=8cm]{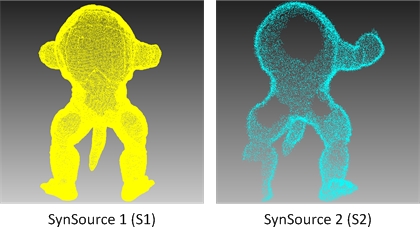}
	\caption{Samples of synthetic cross-source datasets.}
	\label{sampleSyn}
\end{figure}

Database C: Synthetic cross-source point clouds. We build the synthetic datasets according to the cross-source properties. Simulating the cross-source problems discussed in Section \ref{introduction}, we build the synthetic datasets in three steps. Step 1: Different density and different viewpoints. We up-sample the original point cloud by adding one point to the gravity center of each triangle of the original surface. We then remove all points whose $z$ coordinate is less than 0 in the upsampling point cloud, and obtain view 1 as S1. The coordinate system is rotated $60^{\circ}$ relative to the $y$ axis and down-samples every 3 points. We obtain view 2 by removing all the $z\le0$ points. Step 2: Missed point cloud construction. Starting from view 2, we randomly delete ten parts in the plane to simulate a VSFM point cloud. Step 3: Rigid transformation. A random scale of 3 to 5, a random rotation matrix in the $x,y,z$ axis of $30^\circ$ to $60^\circ$, and a random translation in the $z$ axis of 0 to 50\% of the largest point-point distance are added to view 2. Step 4: Construction of noise and outliers. 40DB of noise is added to the original view 2 point cloud. The outliers are constructed by down-sampling the original view 2 to 30\% and adding random offset\footnote{offtset ranges from 0 to 1\% of the largest point-point distance} to the coordinate of the down-sampled point cloud. The noise and outliers are combined to form the final point cloud S2.  The S1 and S2 point clouds are simulating cross-source point clouds which perceive the cross-source problems. Ten cross-source datasets are synthesized using Stanford 3D objects\footnote{http://graphics.stanford.edu/data/3Dscanrep/}. Figure \ref{sampleSyn} shows one sample of the synthetic datasets.

We first compute the radius of the point clouds for parameter setting by $radius=max||P_i-\overline P||_2$, where $\overline{P}=(\sum\limits_{i=1}^{N}{P_i})/N$ is the centroid point of the point cloud. To retain the same density and the same cross-source point cloud structure, we set the radius of the super voxels as 1\% of the point cloud radius for both the KinectFusion and SFM point clouds. For the proposed method, we first compute the transformation matrix on macro and micro structures and then use the transformation matrix to perform transformation on the original cross-source point cloud. 

\subsection{Experiments on same-source point cloud datasets}

\begin{figure*}[ht]
	\centering
	\includegraphics[height=3.0cm,width=150mm]{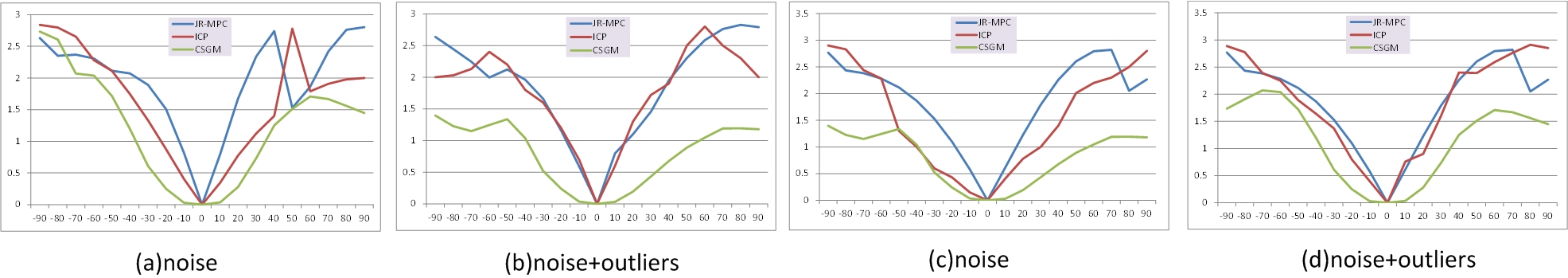}
	\caption{RMSE as a function of the overlap (rotation angle) when two point sets are
		registered (SNR=20dB, 30\% outliers) (a),(b) "Armadillo" (c), (d) "Lucy".}
	\label{f8}
\end{figure*}

We use the root-mean-square error (RMSE) of the rotation parameters for the registration error since translation estimation is not challenging. We select "Armadillo" and "Bunny" with $30^{\circ}$ and $45^{\circ}$ respectively(SNR = 10db and 20\% outliers).

\begin{table}[ht]
	\begin{center}
		\caption{RSME results of the JR-MPC, ICP and CSGM.}
		\label{t1}
		\begin{tabular}{llll}
			\hline\noalign{\smallskip}
			RSME-D & JR-MPC & ICP & CSGM\\
			\noalign{\smallskip}
			\hline
			\noalign{\smallskip}
			Armadillo & {1.456} & 1.725 & 0.508\\
			Bunny & {1.789} & 2.022 & 1.792\\
			\hline
		\end{tabular}
	\end{center}
\end{table}

\begin{figure*}[ht]
	\centering
	\includegraphics[height=8cm,width=180mm]{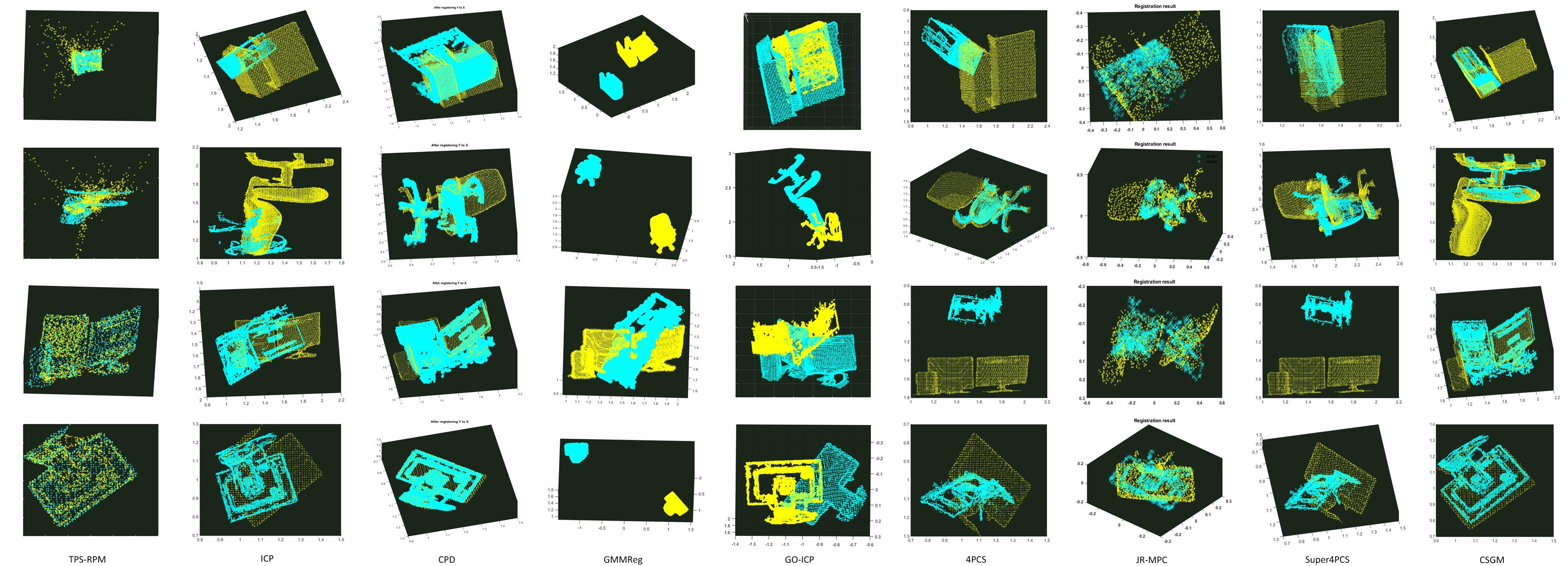}
	\caption{Cross-source point cloud registration results on Database A.}
	\label{f9}
\end{figure*}

Extensive evaluation and comparison of registration methods has been conducted by JR-MPC on same-source databases. We only run JR-MPC, ICP and the proposed method(CSGM) on the same-source database. Table \ref{t1} shows the quantitative comparison results. Note that ICP is more affected by the presence of outliers as a result of the one-to-one correspondence and incurs a higher rate of error. JR-MPC demonstrates similar performance to the proposed method, because GMM models perform well when the overlapping areas do not have a significant amount of missing data or the scale problem. We can see from this experiment that the proposed method is robust to outliers, noise and angle variations on same-source point clouds. The visual results are shown in Figure \ref{f7}.

In addition, we test the robustness of the algorithms in terms of the rotation angle between two point clouds to capture the difference degree of the angles. We register the points under different angles from $-90^{\circ}$ to $90^{\circ}$ and use RMSE to test the performance. The results are shown in Figure \ref{f8} and it can  be seen that the angles have a different effect on the final error. As the proposed method uses a macro and micro structure to describe the point clouds, it shows robustness in dealing with outliers, noise and missing data on same-source database. However, the error increases when the rotation angle increases, similar to other methods. With the increase in the rotation angle, the outliers and the mismatched parts become a larger proportion of each point cloud.

\subsection{Qualitative evaluation on real cross-source point clouds}

As discussed previously, cross-source point clouds have a large variations in density, scale, angle and missing data which makes the already difficult point cloud registration problem even more challenging. To test the ability of our method to register cross-source point clouds and compare with other related methods, we conduct qualitative analysis experiments on four real cross-source datasets: \textit{Twobox, Chair, Threemonitor and Monitor}. To make a thorough comparison, TPS-RPM \cite{rpm}, ICP \cite{icp}, CPD \cite{cpd} , GMMReg \cite{2011gmm}, Go-ICP \cite{goicp}, 4PCS \cite{4pcs}, JP-MPC \cite{jrmpc} and super-4PCS \cite{super4pcs} are selected as our comparison methods. Since many of the selected methods are unable to handle the scale problem, we first normalize the scale difference for ICP, Go-ICP, 4PCS, super-4PCS, TPS-RPM, GMMReg and JP-MPC using our scale normalization method. In our proposed method, scale normalization is an integrated step.

\begin{figure*}[ht]
	\centering
	\includegraphics[height=8cm,width=170mm]{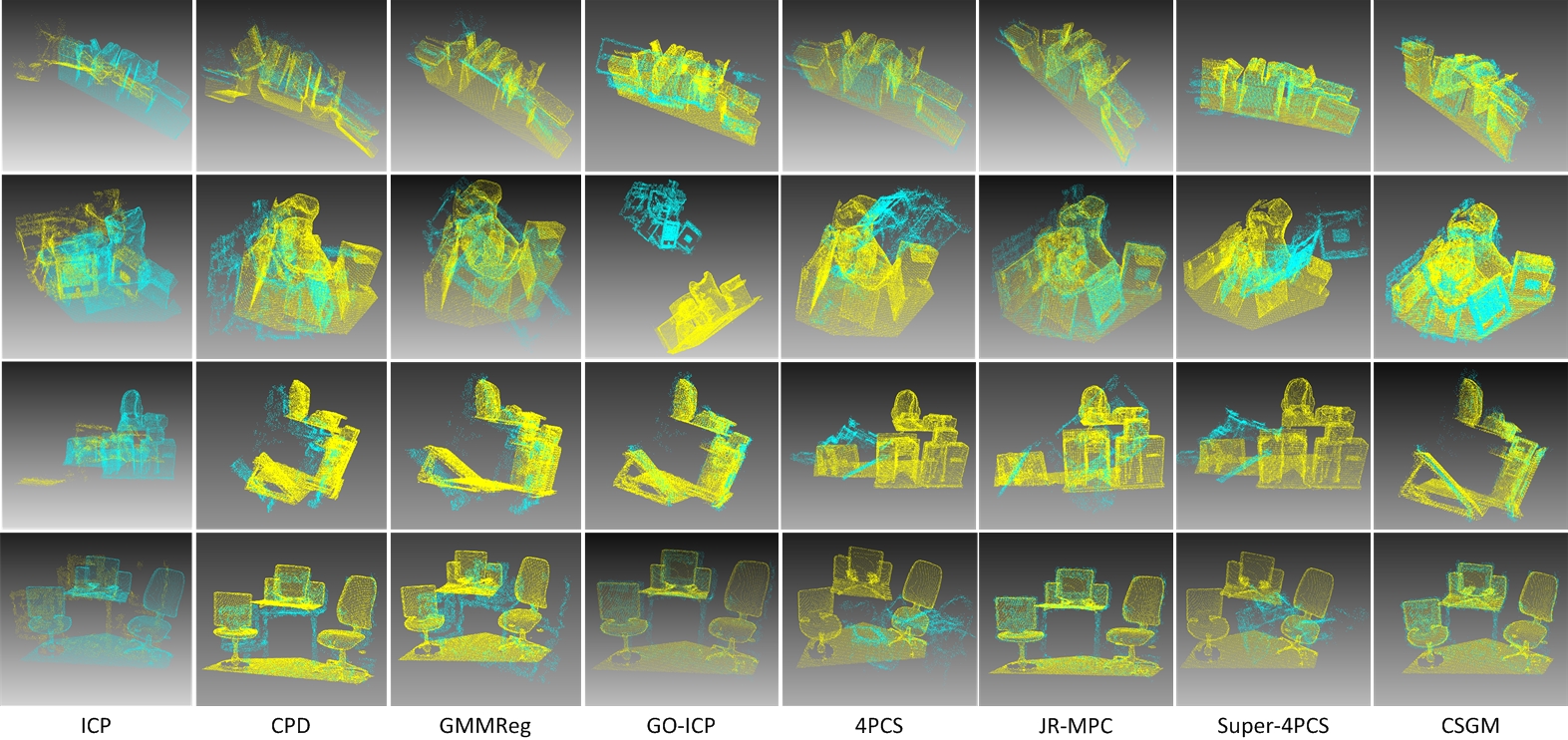}
	\caption{Selected visual effect of cross source point clouds registration results on the Database B. Rows are datasets and columns are methods.}
	\label{visualEffect}
\end{figure*}

Figure \ref{f9} shows the final registration results which indicate that the proposed  method gives successful registration results, whereas the other methods fail in almost all cases. This is because many of these methods cannot handle scale problem, density problem or missing data. Note that TPS-RPM obtains good result in \textit{Threemonitor} and \textit{Monitor}, but fails in \textit{Twobox} and \textit{Chair}. Also, TPS-RPM is a non-rigid registration method. The proposed method obtains good results in cross-source datasets because it describes the micro and macro structure of point clouds, and uses the new optimization method to obtain correspondence relations.

Note that we do not iteratively conduct enhanced graph matching and outlier detection (RANSAC). We find that when we use the outlier detection method to remove graph nodes, the graph structure in some cases is totally different. As a alternative solution, we use ICP to smoothly refine the graph matching result to obtain a final registration result.

\subsection{Quantitative evaluation on real and synthetic cross-source point clouds}

To test the ability of the proposed method, we conduct quantitative evaluation on real and synthetic cross-source databases.

We first conduct quantitative evaluation on Databases B which contains real cross-source point clouds. We compare it in the quantitative evaluation experiments with methods that deal with rigid registration. Based on our knowledge, we compare our proposed method with ICP \cite{icp}, GO-ICP \cite{goicp}, GMMReg \cite{2011gmm}, JP-MPC \cite{jrmpc}, CPD \cite{cpd} and 4PCS \cite{4pcs} and super-4PCS \cite{super4pcs} on a cross-source database. 

\begin{figure}[ht]
	\centering
	\includegraphics[height=6cm,width=8cm]{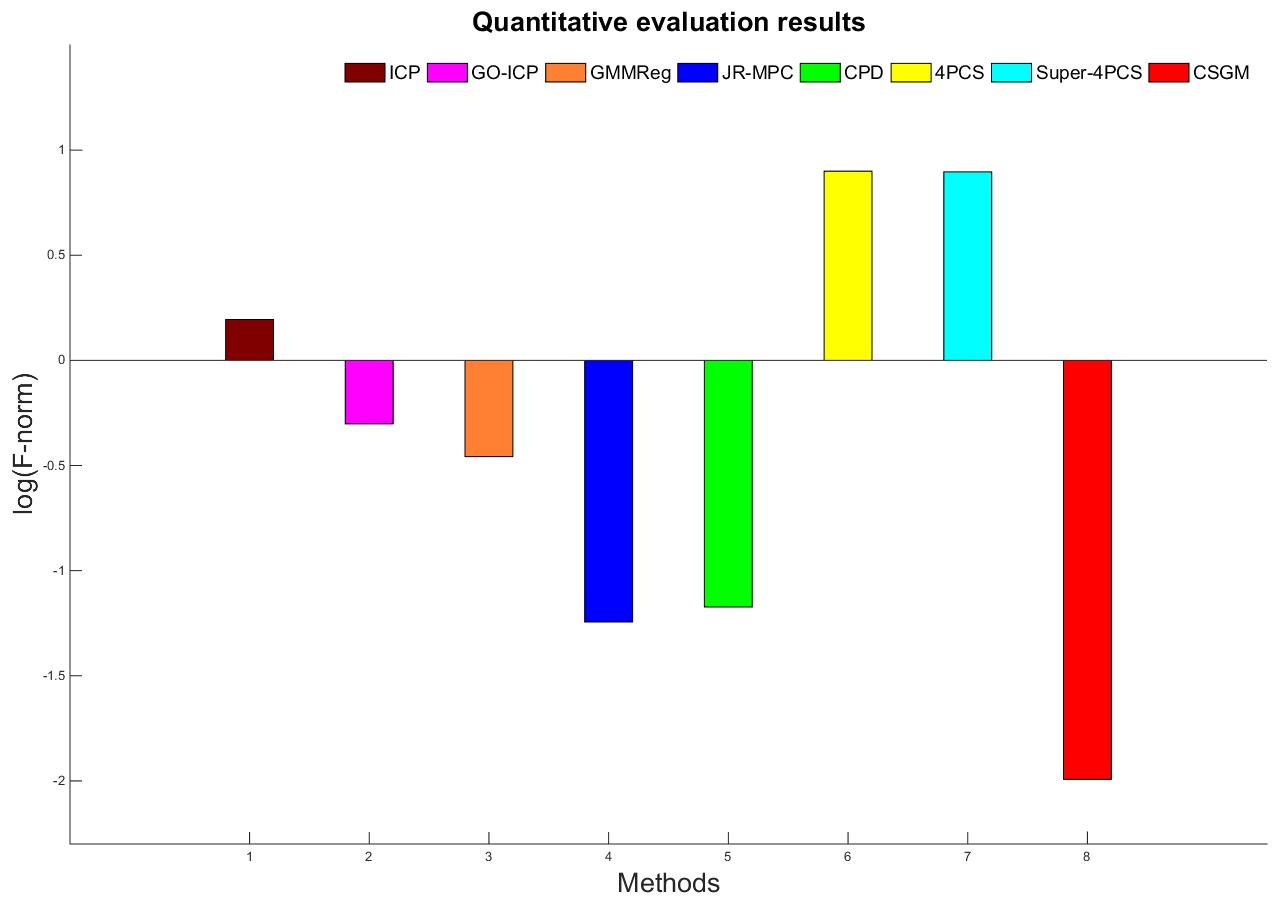}
	\caption{Quantitative evaluation results of mean F-norm between transformation matrices on Database B.}
	\label{Quantative○}
\end{figure}

Many rigid methods are unable to handle the scale problem. To make a fair comparison, scale normalization is performed before running these methods except for CPD which estimates scale internally. The transformation matrix for each comparison method is then computed and these matrices are used for quantitative evaluation. In this experiment, the matrices are all transformed from VSFM point clouds to KinectFusion point clouds. The VSFM point clouds are initially performed by using new computed and ground truth transformation matrices. These transformed VSFM point clouds are then compared with the ground truth transformed point clouds. As in \cite{jrmpc}, we compare the Frobenius Norm (F-norm) between the newly computed matrices and the ground truth transformation matrices. To obtain a better visual representation of comparison results, we use $log(RSME)$ as the final performance value. The smaller the value, the better performance of the algorithm. We also compute the mean of the F-norm of all 13 datasets for each method and the results are shown in Figure \ref{Quantative○}.

\begin{figure*}[ht]
	\centering
	\includegraphics[height=8cm,width=170mm]{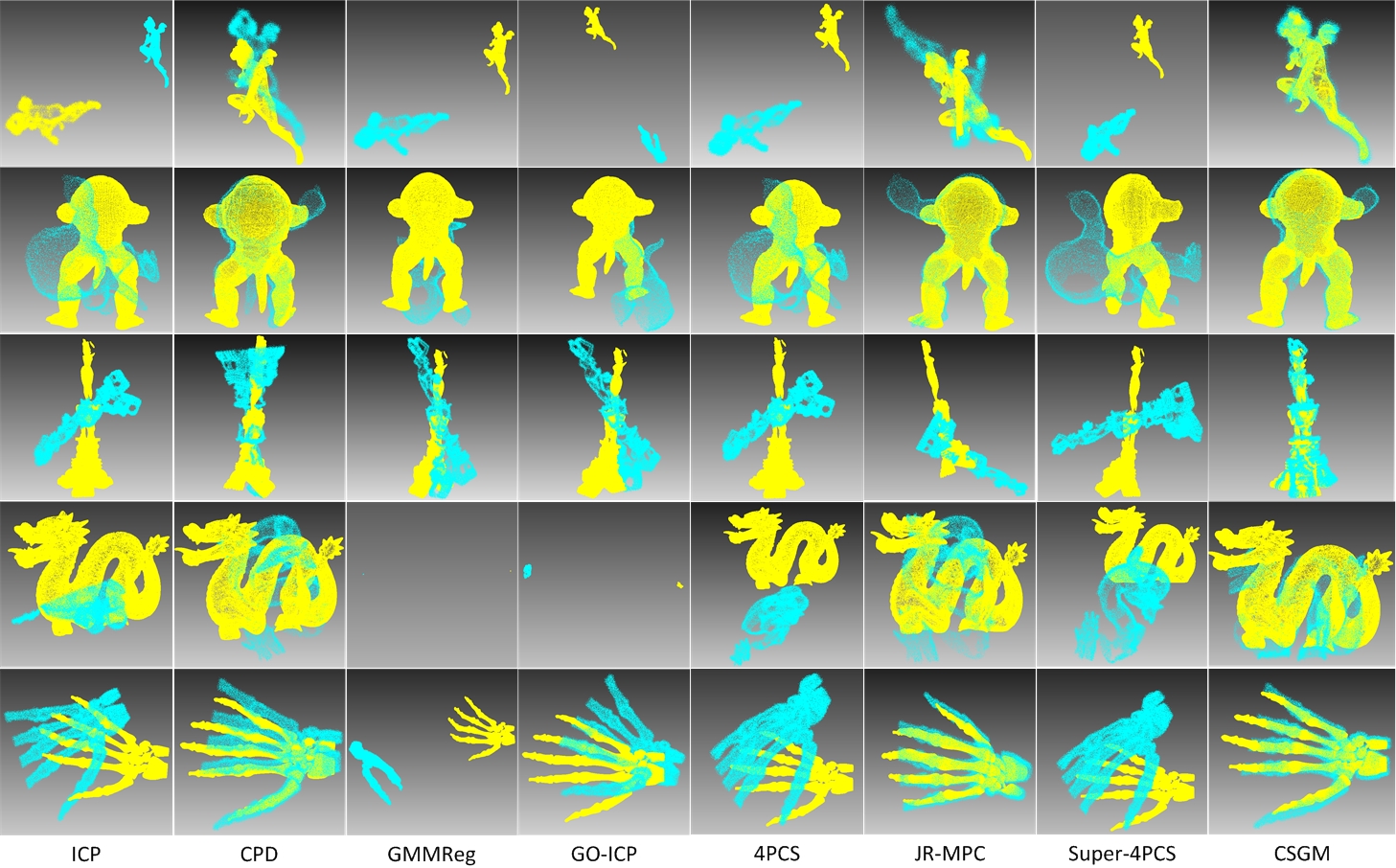}
	\caption{Visual effect of registration results on Database C. Rows are different datasets and columns are methods.}
	\label{visualEffectSyn}
\end{figure*}

Figure \ref{Quantative○} shows the quantitative evaluation results. It illustrates the 4PCS and Super-4PCS obtain worst results, and ICP  follows. It is because the point-point level strategy shows poor ability in cross-source problems. The GMMReg, JR-MPC and CPD show more robust and higher accuracy than other comparison methods; to some extent, they demonstrate the advantage of using the statistical property. The proposed CSGM method obtains the highest accuracy on all dataset. This is because we use the macro structure  to globally register two point clouds with little attention to the detail, and use the micro structure to accurately register the two point clouds. We also use RANSAC and ICP to further improve the accuracy and robustness.

Figure \ref{visualEffect} shows several sample visual results of these methods. The results show that the proposed CSGM  clearly achieves better results than the other methods. Go-ICP and JR-MPC obtain similar results to the proposed CSGM in the fourth row dataset. Because of the BnB strategy in Go-ICP and the generative strategy in JR-MPC, good results are obtained if the scale normalizes very well and no large data are missing. If these conditions do not exist, these methods will completely fail. In the first two rows of Figure \ref{visualEffect}, for example, these methods show the results of that failure. However, the proposed CSGM achieves robust and accurate registration results in all cross-source datasets.

The proposed method is also compared on Database C which consists of synthetic cross-source point clouds. Transformation relation is estimated by the compared methods and the proposed method from view 2 to view 1 point cloud. The computed and ground truth transformation matrix are then utilized to transform the synthetic point cloud. The RSME error is computed according to the statistical distance of these two transformed point clouds. Also, we compare the F-norm of the error of difference between transformation matrices.

\begin{figure}[ht]
	\centering
	\includegraphics[height=4.5cm,width=8cm]{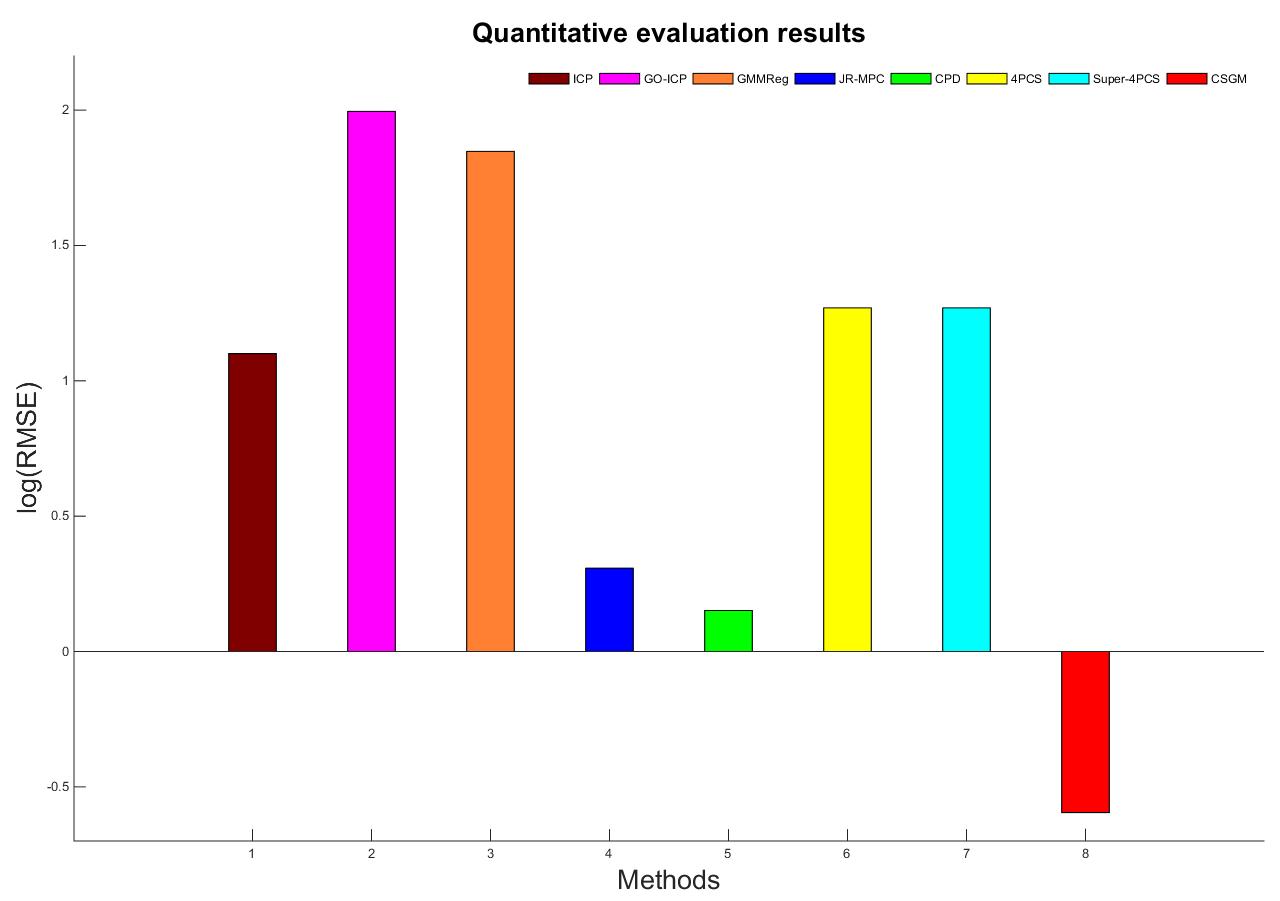}
	\caption{Quantitative evaluation results of RMSE on Database C.}
	\label{SyntheticRMSE}
\end{figure}

\begin{figure}[ht]
	\centering
	\includegraphics[height=4.5cm,width=8cm]{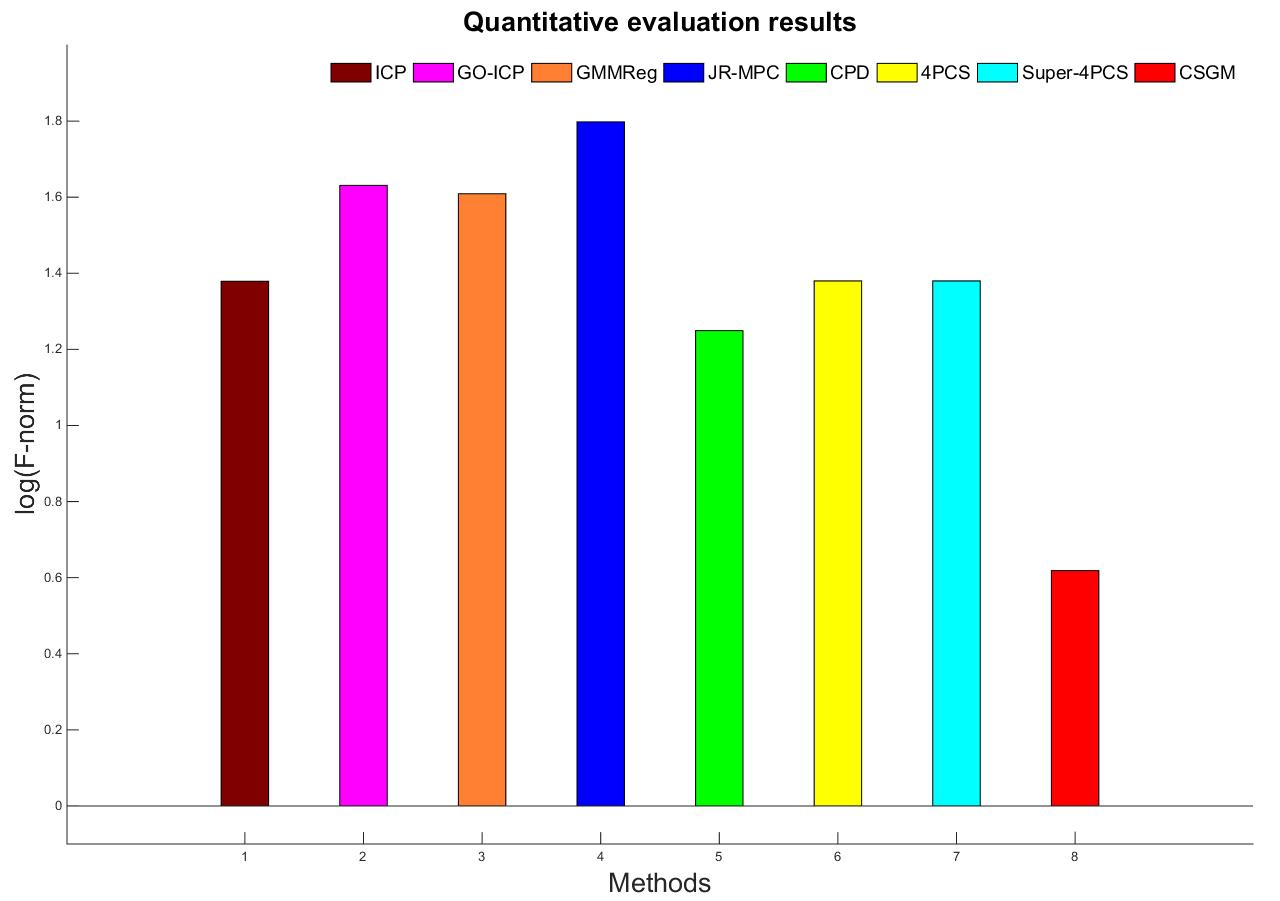}
	\caption{Quantitative evaluation results of F-norm on Database C.}
	\label{SyntheticFnorm}
\end{figure}

Figure \ref{SyntheticRMSE} shows the evaluation results of mean RMSE and Figure \ref{SyntheticFnorm} shows the evaluation results of mean F-norm of the computed transformation matrix and the ground-truth transformation matrix on whole ten sets of Database C. The results show that our method achieves accurate registration results which are better than the other methods.  Figure \ref{visualEffectSyn} illustrates the visual effects of the Synthetic evaluation. The results show that the proposed CSGM obtains robust and visually correct registration results which are clearly better than those of the compared methods. Some of the comparison methods are even failed because the cross-source problem are really great challenge to these methods.


\section{Conclusion}
\label{conclusion}
In this paper, we proposed a new registration pipeline to deal with the cross-source point cloud registration problem using four novelty components. A scale normalization method was first proposed to eliminate the scale problem.  Secondly, a micro and macro structure concept was proposed to describe the point clouds, and a new graph construction method was used to combine these structures. Thirdly, an optimization method was proposed to solve the problem. Lastly, a registration pipeline was proposed which combines the initial correspondence from graph matching and refinement using RANSAC and ICP.



%

%

\section*{Acknowledgment}

The authors would like to thank the Nokia Corporation for their help and acknowledge the useful discussions with colleagues in GBDTC. This work is partially supported by Nokia research funding (MM12030846235).

\ifCLASSOPTIONcaptionsoff
  \newpage
\fi



%

\bibliographystyle{IEEEtran}
\bibliography{BIB}

%
%
%
%
%
%
%
%
%
%

\end{document}